\documentclass[journal]{IEEEtran}
\usepackage{cite}
\usepackage[pdftex]{graphicx}
\usepackage{amsmath}
\usepackage{amsfonts}
\usepackage[caption=false,font=footnotesize]{subfig}
\usepackage{tabu}
\usepackage{hyperref}
\usepackage{url}
\usepackage{placeins} 

\newcommand{\gomez}{G\'{o}mez}
\newcommand{\bt}[1]{\textbf{#1}}
\newcommand{\etal}{\textit{et al.}}
\newcommand{\todo}[1]{\iffalse #1 \fi}

\begin{document}
\title{MSMatch: Semi-Supervised Multispectral Scene Classification with Few Labels}

\author{\IEEEauthorblockN{Pablo \gomez\textsuperscript{*} and
Gabriele Meoni\textsuperscript{*}}
\thanks{P. \gomez{} and Gabriele Meoni are with the Advanced Concepts Team, European Space Agency, Noordwijk, The Netherlands e-mail: pablo.gomez@esa.int, gabriele.meoni@esa.int}
\thanks{Manuscript received ...; revised .... }}


\maketitle

\begin{abstract}
Supervised learning techniques are at the center of many tasks in remote sensing. Unfortunately, these methods, especially recent deep learning methods, often require large amounts of labeled data for training. Even though satellites acquire large amounts of data, labeling the data is often tedious, expensive and requires expert knowledge. Hence, improved methods that require fewer labeled samples are needed. \\
We present MSMatch, the first semi-supervised learning approach competitive with supervised methods on scene classification on the EuroSAT and UC Merced Land Use benchmark datasets. We test both RGB and multispectral images of EuroSAT and perform various ablation studies to identify the critical parts of the model. \\
The trained neural network achieves state-of-the-art results on EuroSAT with an accuracy that is up to 19.76\% better than previous methods depending on the number of labeled training examples. With just five labeled examples per class, we reach 94.53\%  and 95.86\% accuracy on the EuroSAT RGB and multispectral datasets, respectively. On the UC Merced Land Use dataset, we outperform previous works by up to 5.59\%  and reach 90.71\% with five labeled examples.
Our results show that MSMatch is capable of greatly reducing the requirements for labeled data. It translates well to multispectral data and should enable various applications that are currently infeasible due to a lack of labeled data. We provide the source code of MSMatch online to enable easy reproduction and quick adoption.
\end{abstract}

\begin{IEEEkeywords}
Deep learning, multispectral image classification, scene classification, semi-supervised learning, 
\end{IEEEkeywords}

\begingroup\renewcommand\thefootnote{*}
\footnotetext{Equal contribution}
\endgroup

\IEEEpeerreviewmaketitle

\section{Introduction}
\IEEEPARstart{T}{he} last decade has seen a momentous increase in the availability of remote sensing data, thus enhancing the need for efficient image processing and analysis methods using deep learning \cite{zhu2017}. The former is driven by continuously decreasing launch costs, especially for so-called Smallsats ($< 500$kg). As Wekerle \etal{} \cite{wekerle2017} describe, less than 40 Smallsats were launched per year between 2000 and 2012 but over a hundred in 2013 and almost 200 in 2014. Since then the numbers have been increasing with over 300 launches in 2018 and 2017 \cite{puteaux2019}. Many of these are imaging satellites serving either commercial purposes \cite{popkin2017} or related to earth observation programs, such as the European Space Agency's Copernicus program \cite{aschbacher2017,wekerle2017}. This has led to an increase in the availability of large datasets. \\
Concurrently, image processing and analysis have improved dramatically with the advent of deep learning methods \cite{alom2018}. As a consequence, there is a large corpus of research describing successful applications of deep learning methods to remote sensing data \cite{ball2017,chen2016,ma2019,zhang2016,zhu2017}. However, training deep neural networks usually requires large amounts of labeled samples, where the expected solution has been manually annotated by experts \cite{chen2017hyperspectral}. This is in particular tedious for imaging modalities such as radar data or multispectral (MS) imaging data, which is not as easily labeled by humans as, e.g., RGB imaging data. \\
One way to alleviate these issues is the application of so-called semi-supervised learning (SSL) techniques. These aim to train machine learning methods, e.g. neural networks, while providing only a small set of labeled training samples and a typically larger corpus of unlabeled training samples. This has recently garnered a lot of attention in the remote sensing community \cite{liu2017,wu2017,zhang2020,roy2018,tao2020}. In the last two years, the state-of-the-art in SSL has advanced significantly\footnote{see \url{ https://paperswithcode.com/task/semi-supervised-image-classification} (Accessed 17.03.2021)} to a point, where the proposed methods are virtually competitive with fully supervised approaches \cite{berthelot2019,berthelot2019remixmatch,kurakin2020}. The recent advances in SSL approaches bear the promise to save large amounts of time and cost that would be required for manual labeling. \\
In this work, we propose MSMatch, a novel approach that builds on recent advances \cite{kurakin2020} together with recent neural network architectures (so-called EfficientNets \cite{tan2019}) to tackle the problem of land scene classification, i.e. correctly identifying land use or land cover of satellite or airborne images. This is an active research problem with a broad range of research focusing on it \cite{cheng2017,cheng2020,gu2019}.
We compare with previous methods on two datasets, the EuroSAT benchmark dataset \cite{helber2018,helber2019} collected by the Sentinel-2A satellite and the aerial UC Merced Land Use (UCM) dataset \cite{yang2010}. The EuroSAT dataset also includes MS data, which is commonly used for tasks related to vegetation mapping \cite{chen2016}. \\
In summary, the main contributions of this work are:

\begin{itemize}
\item First SSL approach that is competitive with supervised methods on scene classification on EuroSAT and UCM

\item MSMatch can reach 94.53\% and 95.86\% accuracy on EuroSAT RGB and MS and 90.71 \% on UCM, respectively, with only five labels per class. This greatly reduces the need for labeled data.
 
\item Extending the applicability of recent advances in neural network architectures and SSL approaches to MS data and remote sensing
  
\item Analyses of critical components of the proposed pipeline by performing ablation studies and identification of heterogeneous properties in the classes in EuroSAT and UCM

\end{itemize}

\section{Related Work}
The need for a large amount of labeled training data is one of the most significant bottlenecks in bringing deep learning approaches to practical applications \cite{ma2019}. For satellite imaging data this problem is particularly aggravated as satellites have greatly varying sensors and applications, which makes a transfer between a model trained on data from one application or sensor to another challenging \cite{furano2020towards}. Hence, SSL is of particular interest in remote sensing. In the following, we describe relevant works that applied SSL to scene classification problems. Further, we point out the works that led to significant improvements in SSL in the broader machine learning community in the last years. 
\subsection{Semi-supervised learning for scene classification}
There are several datasets that have been established as benchmark datasets for scene classification. Some of the most commonly used ones are EuroSAT, UCM and the Aerial Image Dataset (AID) \cite{xia2017}. Both, UCM and AID, use aerial imaging data and provide, respectively, 2100 and 10000 images for a classification of 21 and 30 classes. EuroSAT provides 27000 images of 10 classes. Aside from the SSL works mentioned here, there is also a multitude of studies using supervised methods for these datasets (e.g. \cite{zhang2019,xie2019,pires2020}). \\
In terms of SSL approaches, there have been several interesting approaches: Guo \etal{} \cite{guo2020} trained a generative adversarial network (GAN) that performs particularly well with few labels on UCM and on EuroSAT. They achieved between 57\% and 90\% accuracy (4.76 to 80 labels per class) on UCM and between 77\% to 94\% on EuroSAT (10 to 216 labels per class). Han \etal{} \cite{han2018} used self-labeling to achieve even better results on UCM reaching 91\% to 95\% using comparatively more labels. They report similarly good results with comparatively large label counts (10\% of the whole data) on AID. Dai \etal{} \cite{dai2019} used ensemble learning and residual networks with much fewer labels on UCM and on AID reaching  85\% on UCM and between 72\% and 85\% on AID. Similiarly, Gu \& Angelov proposed a deep rule-based classifier using just one to ten labels per class and achieving between 57\% and 80\% on UCM \cite{gu2018semi}. \\ 
A self-supervised learning paradigm was suggested by Tao \etal{} \cite{tao2020} for AID and EuroSAT which obtained between 77\% to 81\% on AID and 76\% to 85\% on EuroSAT. Another GAN-based approach has been suggested in the work of Roy \etal{} \cite{roy2018}, who applied it to EuroSAT, but other approaches have already accomplished better results, such as the work by Zhang and Yang \cite{zhang2020}, who utilized the EuroSAT MS data (97\% accuracy) - however with 300 labels per class. Yamashkin \etal{} \cite{yamashkin2020} also suggested an SSL approach where they extended dataset, but their results are not competitive. Thus, reaching high accuracy over 90\% on any of the datasets usually still requires a larger amount of labels (80 per class or more). Low label regimes with e.g. five labels per class typically reach only about 70\% to 80\% accuracy.


\subsection{Recent advances in semi-supervised learning}

Semi-supervised learning is a field that has been gaining a lot of attention in recent years. In particular, the last two years have seen several methods being published that led to unprecedented results and, for the first time, achieved results that are competitive with supervised methods trained on significantly more data. For example, the accuracy on the popular CIFAR-10 dataset \cite{krizhevsky2009} for training with just 250 labels has improved from 47\% to 95\% \footnote{see \url{ https://paperswithcode.com/sota/semi-supervised-image-classification-on-cifar-6} (Accessed 11.03.2021)} from 2016 to 2020. Many of these improvements rely on smart data augmentation strategies such as RandAugment \cite{cubuk2020} or AutoAugment \cite{cubuk2018}. The most significant improvements were made in 2019 in two works describing the so-called \textit{MixMatch} \cite{berthelot2019} method and \textit{Unsupervised Data Augmentation} \cite{xie2019UDA}. The former improved the state-of-the-art by over 20\% and the latter pushed the accuracy above 90\%. In a series of follow-up works \cite{berthelot2019remixmatch,nair2019} results were further improved until the current state-of-art method was introduced in 2020. Utilizing the ideas of pseudo-labeling and consistency regularization by Bachman \etal{} \cite{bachman2014}, FixMatch \cite{kurakin2020} achieved state-of-the-art result on four benchmark datasets including almost 95\% accuracy on CIFAR-10 with 250 labels. This is comparable to the performance of a supervised approach for the utilized network architecture. Furthermore, using just four labels per class, they still achieved 89\% accuracy.
To the authors' knowledge, none of the mentioned works have found their way into the remote sensing community yet. Hence, this work aims to build on these recent advances to achieve state-of-art results.

\begin{figure*}[ht]
\centering
\includegraphics[width=\textwidth]{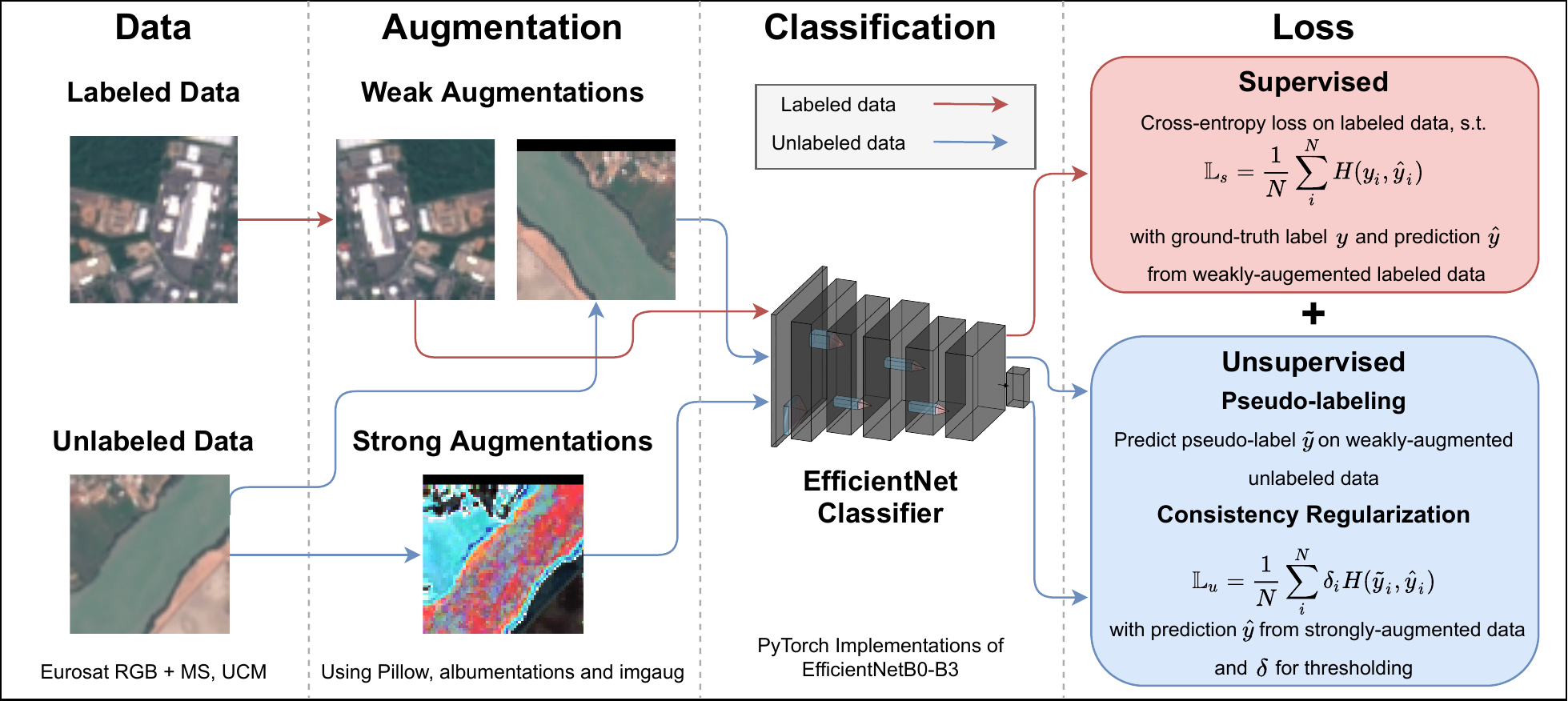}
\caption{Overview of the processing pipeline of MSMatch.}
\label{fig: msmatchPipe}
\end{figure*}

\begin{figure}[ht]
\centering
\includegraphics[scale=1.3]{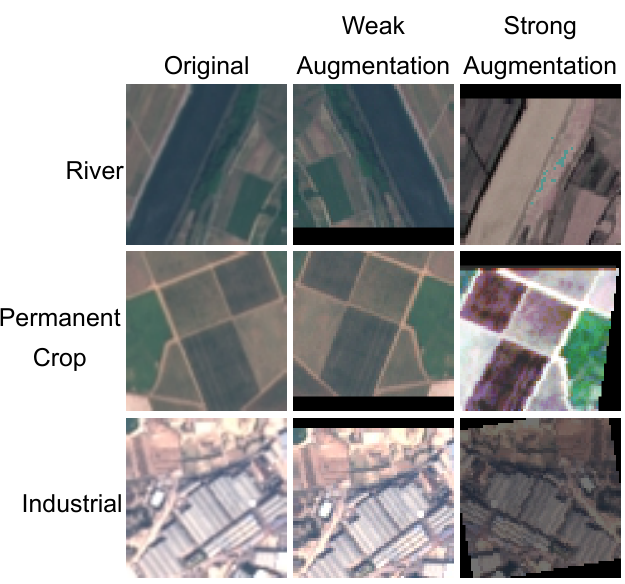}
\caption{Examples of weak and strong augmentation applied to different EuroSAT images. In this picture, strong augmentation effects applied to the class \textit{River} are \textit{AutoContrast}, \textit{Color}, and \textit{Solarize}; the class \textit{Permanent Crop} is augmented through \textit{Sharpness}, \textit{ShearX},and \textit{Equalize}; finally, the class \textit{Industrial} is augmented through \textit{Posterize}, \textit{Rotate}, and \textit{Brightness}.}
\label{fig: augmentation}
\end{figure}

\section{Methods}
This section will introduce the network architecture, SSL method and setup of the training. 

\subsection{EfficientNet}
EfficientNets \cite{tan2019} have become the go-to neural network architecture for many applications. They achieve state-of-art results, especially in terms of efficiency as EfficientNets use comparatively few parameters in relation to achieved performance. The architecture was conceived using neural architecture search \cite{elsken2019neural}, a method where the neural network architecture itself is optimized. Tan and Le propose several versions of EfficientNets called B0 to B7 with increasing numbers of parameters and performance. Thereby, it is possible to choose a suitable trade-off that keeps memory and computational requirements manageable at a sufficient model complexity.
Note, however, that EfficientNets have not seen broad adoption in the remote sensing community, yet. None of the mentioned prior SSL works utilized them. We utilize them for MSMatch given their excellent performance and low memory footprint. We relied on an open-source implementation utilizing \textit{PyTorch}.\footnote{\url{https://github.com/lukemelas/EfficientNet-PyTorch} (Accessed 11.03.2021)}
For MS images the first network layer received all available bands as normalized input channels individually. Thus, the number of network parameters is almost identical for MS and RGB images and performance differences ought to be largely due to contained information.

\subsection{FixMatch}

There are two central ideas behind the effectiveness of the FixMatch approach, pseudo-labeling and consistency regularization \cite{bachman2014,kurakin2020}. Pseudo-labeling refers to practice of using the model (or another model) to automatically label otherwise unlabeled data. The second idea, consistency regularization, refers to the concept that the model should predict the same output for similar inputs. FixMatch training consists of a supervised and unsupervised loss. While the supervised loss is a common cross-entropy loss, the unsupervised loss incorporates both pseudo-labeling and consistency regularization.
This is achieved by creating two different augmentations of the same image, a so-called weak and a strong one. As depicted in Figure \ref{fig: msmatchPipe}, the weakly augmented image is used to create a pseudo-label for the image. Consistency regularization is then employed by computing a cross-entropy loss between a pseudo-label on the weakly augmented image and the model's classification of the strongly augmented image.
Thus, the supervised loss is
\begin{equation}
    \mathbb{L}_s = \frac{1}{N}\sum_i^N  H(y_i, \hat{y}_i),
\end{equation}
where $H$ is a cross-entropy loss, $y_i$ the ground-truth label and $\hat{y}_i$ the network's prediction on the weakly augmented Image. The unsupervised loss is
\begin{equation}
    \mathbb{L}_u = \frac{1}{N}\sum_i^N \delta_i H(\tilde{y}_i, \hat{y}_i)
\end{equation}
where $\tilde{y}_i$ is the pseudo-label on an unlabeled, weakly augmented image and $\delta_i$ is $1$ if $\tilde{y}_i \geq 0.95$ and $0$ otherwise.
The total loss from the supervised loss $\mathbb{L}_s$ and unsupervised loss $\mathbb{L}_u$ is then obtained as $\mathbb{L} = \mathbb{L}_s + \mathbb{L}_u$. \\
For the implementation of FixMatch we adapted an open-source \textit{PyTorch} implementation\footnote{\url{https://github.com/LeeDoYup/FixMatch-pytorch} (Accessed 11.03.2021)}. To the authors' knowledge this is also the first work applying FixMatch to MS images.

\subsection{Augmentation}
Data augmentation is frequently used to help neural networks generalize better to unseen data and to increase the richness of the utilized training data \cite{cubuk2018,cubuk2020}. During the FixMatch training process the augmentation is critical to ensure that the little amount of available labeled data is exploited optimally using the weak augmentations, and to aid generalization to unseen data using the strong augmentations. \\
For the weak augmentation, we only utilized horizontal flips and image translations by up to 12.5\%.
Note, that Kurakin \etal{} \cite{kurakin2020} describe in their work that, e.g., they tried to harness stronger augmentations for the labeled data but experienced training divergence. We encountered similar issues utilizing, e.g., image crops. For the strong augmentation of the RGB images, several methods from the Python library \textit{Pillow} were applied. For the strong augmentation of the MS data, a slightly reduced set (due to a lack of implementations for more than three image channels) were applied utilizing the \textit{albumentations} Python module \cite{buslaev2020}. 
Exemplary weak and strong augmentations of RGB EuroSAT images are depicted in Figure \ref{fig: augmentation}. A full overview of the applied augmentations can be seen in Table \ref{table:aug}.
\begin{table}
\caption{Overview of applied strong augmentations. RGB images were augmented using the Python library \textit{Pillow}, MS images with the Python library \textit{albumentations}. Detailed parameter overviews are given in the libraries' respective documentation. \label{table:aug}}
\begin{center}
 {\tabulinesep=0.5mm
\setlength\tabcolsep{2pt}
\begin{tabu} {|X[0.5]|| X[0.25c] | X[0.25c] | X[1c]|}
 \hline
  Augmentation & RGB & MS & Parameterrange \\
 \hline
AutoContrast    & x & &     -                 \\ \hline
Brightness      & x & &     [0.05, 0.95]      \\ \hline
Color           & x & &     [0.05, 0.95]      \\ \hline
Contrast        & x & x &   [0.05, 0.95]      \\ \hline
Equalize        & x & x &   -                 \\ \hline
Posterize       & x & x &   [4, 8]            \\ \hline
Rotate          & x & x &   [-30, 30]         \\ \hline
Sharpness       & x & x &   [0.05, 0.95]      \\ \hline
ShearX          & x & x &   [-0.3, 0.3]       \\ \hline
ShearY          & x & x &   [-0.3, 0.3]       \\ \hline
Solarize        & x & x &   [0, 256]          \\ \hline
TranslateX      & x & x &   [-0.3, 0.3]       \\ \hline
TranslateY      & x & x &   [-0.3, 0.3]       \\ \hline
\end{tabu}}
\label{table:augmentations}
\end{center}
\end{table}

\subsection{Training}
All models were trained for three different random seeds on NVIDIA RTX 2080 TI graphics cards using \textit{PyTorch} 1.7. The training utilized a stochastic gradient descent optimizer with a Nesterov momentum of 0.9 and different weight decay amounts. A learning rate of 0.03 was used and reduced with cosine annealing. Training batch size was 32 for EuroSAT datasets and 16 for UCM, with one batch containing that many images and additionally four and seven times as many unlabeled ones for UCM and EuroSAT, respectively. The training was run for a total of 500 and 1000 epochs with 1000 iterations each for EuroSAT and UCM, respectively, after which all investigated models had converged. For UCM, the number of epochs was doubled to compensate the smaller batch size.
All images were normalized to the mean and standard deviation of the datasets. The train and test sets were stratified. The test sets for each seed contained 10\% of the data for EuroSAT (2700 images) and 20\% of the data for UCM (420 images). To speed up the training and to reduce the memory footprint, UCM images were downscaled to $224 \times 224$.
For a supervised baseline the unsupervised loss $\mathbb{L}_u$ was set to $0$ to allow a fair and direct comparison. Overall, with this setup, training of one model requires up to 48 hours on a single GPU for EuroSAT. A single run for UCM requires 131 hours on two GPUs. We provide the code for this work open source online\footnote{\url{https://github.com/gomezzz/MSMatch}}.

\section{Results}
We investigate results on two datasets. We report results for both, the RGB and MS, versions of EuroSAT as well as the UCM dataset. Aside from a detailed comparison with previous research depicted in Table \ref{table:main_results} and Table \ref{table:ucm}, we also investigated the impact of the weight decay strength in MSMatch as well as the number of parameters of the utilized EfficientNet. \\

\subsection{Datasets}
Some of the most commonly utilized benchmarks for SSL in remote sensing are UCM and EuroSAT. They allow for a detailed comparison to previous works.

\subsubsection{EuroSAT}
 Given the comparatively large images
, the computational burden of the heavy image augmentation and training the model is larger for datasets like UCM and AID. Further, they provide fewer and only RGB images. Hence, we intensively tested on the EuroSAT dataset, which consists of 27 000 $64 \times 64$ pixel images in RGB and 13-band MS format. The MS bands are between $443$ nm and $2190$ nm, the spatial resolution is up to $10$ meters per pixel depending on the band. Note that especially the infrared bands are well-suited to vegetation identification. The data stem from the Sentinel-2A satellite and are split into ten classes, such as river, forest, permanent crop and similar ones. Some examples of images from the EuroSAT dataset can be seen in Figure \ref{fig: augmentation}. The data were obtained from the authors' \textit{GitHub} respository.\footnote{\url{https://github.com/phelber/EuroSAT} (Accessed 11.03.2021)} 

\subsubsection{UCM}
\begin{figure*}[ht]
\centering
\includegraphics[scale=0.55]{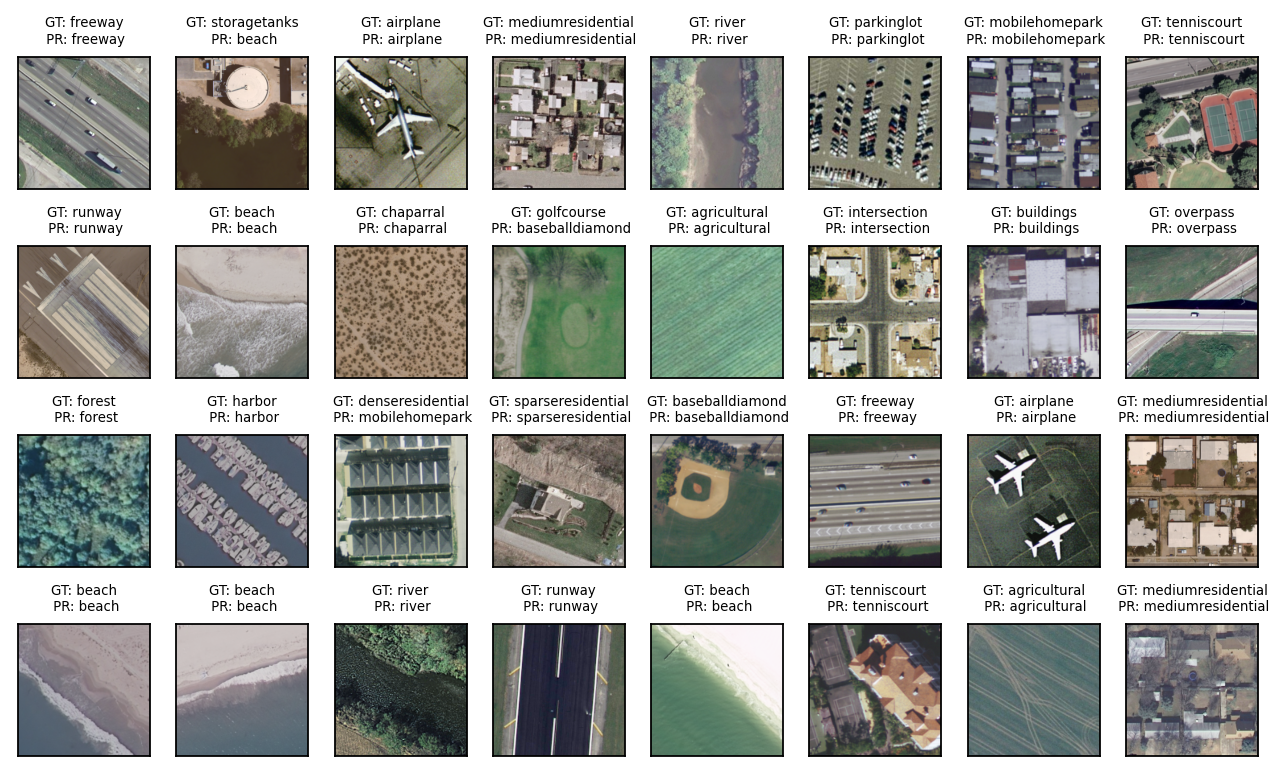}
\caption{Examples of UCM images with ground-truth labels (GT) and predicted classes (PR). Prediction was performed with an EfficientNet-B2 model trained with five images per class on one seed.}
\label{fig: ucm_images}
\end{figure*}

The UCM dataset is arguably the most established land scene classification dataset. It consists of $2100$ images of areas in the USA classified into $21$ classes, such as beach, forest or storage tanks. The original images were taken using aerieal orthoimagery and processed into slices of $256 \times 256$ pixels. Each class is represented with a $100$ images. We display several example images from UCM with the associated labels and predicted classes in Figure \ref{fig: ucm_images}. The data were obtained from the authors' website.\footnote{\url{http://weegee.vision.ucmerced.edu/datasets/landuse.html} (Accessed 13.07.2021)}

\subsection{Number of Labels}
\begin{figure}[ht]
\centering
\includegraphics[scale=0.225]{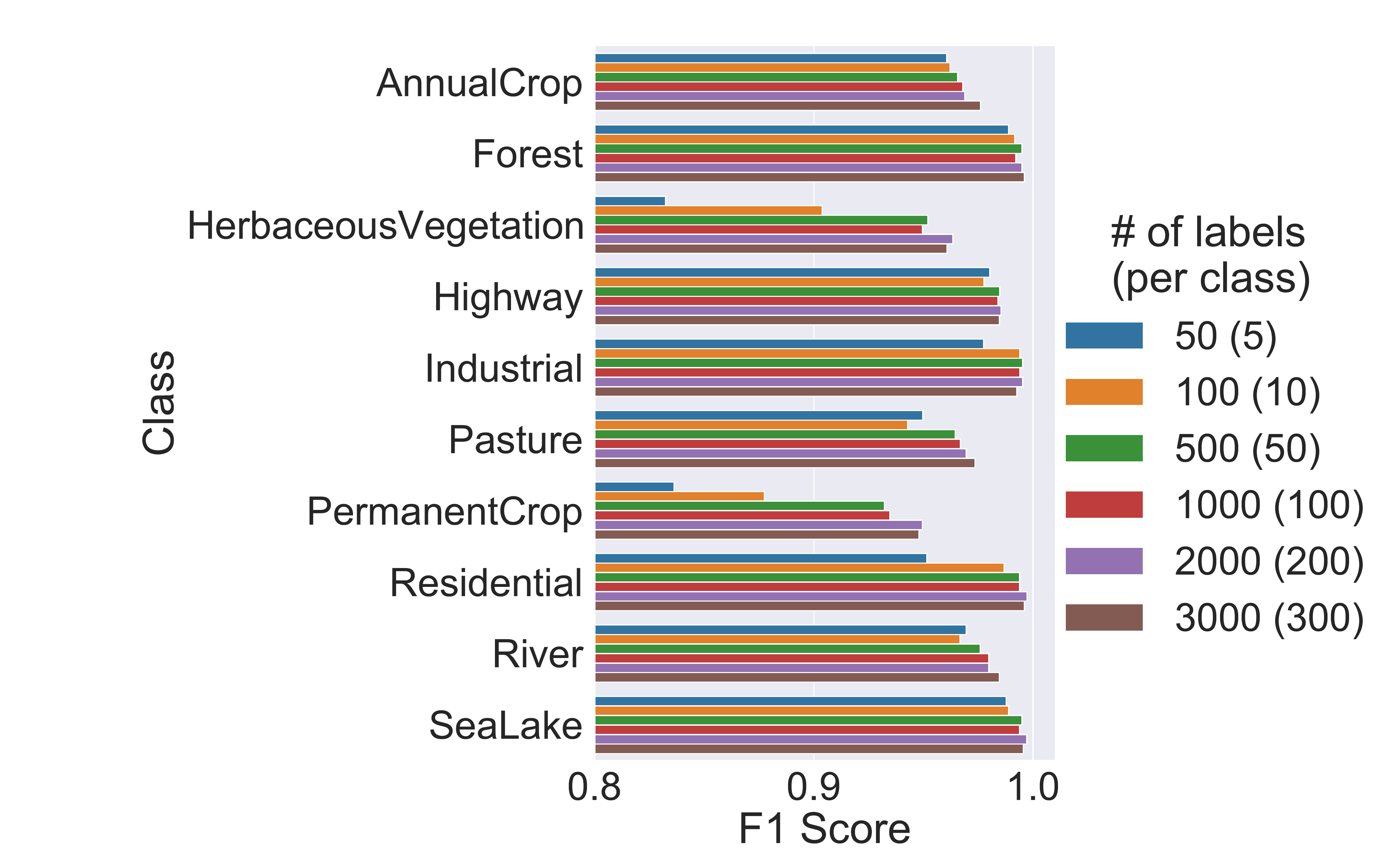}
\caption{Classification F1 scores for models trained with a different amount of labeled samples. RGB data was used. Notably, some classes, such as \textit{PermanentCrop}, seem to require more samples than others, such as \textit{SeaLake}, to reach good scores. Results are averaged over three seeds.}
\label{fig: f_scores_eurosat}
\end{figure}

\begin{figure}[ht]
\centering
\includegraphics[scale=0.22]{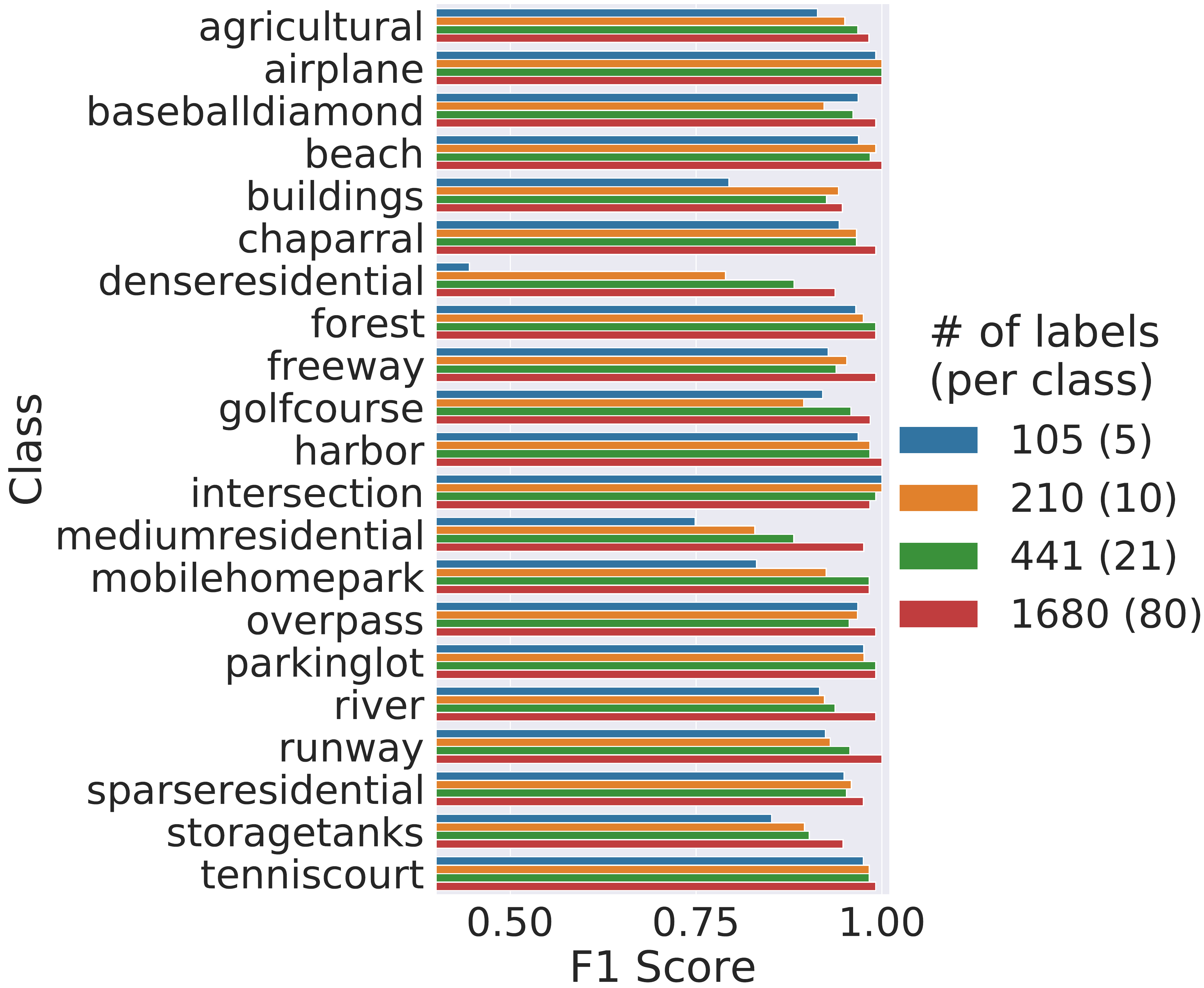}
\caption{Classification F1 scores for models trained with a different amount of labeled samples. Notably, some classes, such as \textit{denseresidential}, seem to require more samples than others, such as \textit{airplane}, to reach good scores. Results are averaged over three seeds.}
\label{fig: f_scores_ucm}
\end{figure}

The main factor for comparing SSL approaches is the number of labeled training examples used for training. For EuroSAT, we tested a large range of different amounts ranging from 50 (five per class) to 3000 (300 per class) labels to ensure comparability with prior research. For UCM, training datasets with 105, 210, 441 and 1680 labels (respectively 5, 10, 21 and 80 labels per class) were investigated. The rest of the (unlabeled) data are used for the unsupervised part of the training.
As seen in Table \ref{table:main_results} and Table \ref{table:ucm}, MSMatch outperforms all prior research by large margins for all tested amounts of available labels. The greatly enhanced accuracy per labeled training sample is especially prominent for the cases with just 50 and 100 labels (five and ten per class). In these cases MSMatch improves on previous methods between 16\% and 20\% on EuroSAT and 2.71\% and 5.59\% on UCM.
Using 1000 (100 per class) labels, which was the most popular amount in prior research on EuroSAT, it improves the state-of-the-art by 7\%.
\\
For EuroSAT, the last three rows in Table \ref{table:main_results} showcase the impact of using MS data and difference if, instead of using our SSL approach, we train an EfficientNet using only the labeled samples and no unlabeled samples. Note, that results on supervised baseline, i.e. training without any unlabeled data, are clearly worse. This demonstrates the effect of the proposed SSL framework. Even for 3000 labels, the SSL method performs significantly better than the baseline. Notably, the MS data improves results even further. The proposed method is hence successfully adapted to MS data.
Due to the high computational demands of training a model UCM, this ablation was not performed on UCM.
Additionally, Figure \ref{fig: f_scores_eurosat} and Figure \ref{fig: f_scores_ucm} show F1 scores for all classes and amount of training labels. Notably, some classes, such as \textit{PermanentCrop} for EuroSAT or \textit{denseresidential} for UCM, seem to require more samples to reach optimal results.

\begin{table}
\caption{Accuracy results comparison on EuroSAT in percent. Works using MS data are marked with an asterisk. MSMatch outperforms all other methods on EuroSAT. The best results per label amount are bold.}
\begin{center}
 {\tabulinesep=1.2mm
\setlength\tabcolsep{2pt}
\begin{tabu} {|X[4]|| X[1] | X[1] | X[1]| X[1] | X[1] | X[1] |}
 \hline
 & \multicolumn{6}{c|}{Number of labels (per class)}\\
  Work & 50 \: (5) & 100 (10) & 500 (50) & 1000 (100) & 2000 (200) & 3000 (300)\\
 \hline
Guo \etal{} 2020 \cite{guo2020}         & -         & 76.79     & -         & 88.72     & 90.66     & -     \\ \hline
Roy \etal{} 2018 \cite{roy2018}        & -         & 68.60     & -         & 86.10     & 89.00     & -     \\ \hline
Tao \etal{} 2020 \cite{tao2020}        & 76.10     & -         & -         & -         & -         & -     \\ \hline
Zhang \& Yang 2020*  \cite{zhang2020}   & -         & 80.12     & 89.01     & 91.11     & -         & 96.67 \\ \hline
Supervised Baseline    & 40.75     &   54.63   &  77.99    &  87.41    & 91.74     & 93.94 \\ \hline
Ours (RGB)              & 94.53& 96.04 & 97.62& 97.63     & 98.07    & 98.14 \\ \hline
Ours (MS)* & \bf{95.86}  & \bf{96.63} & \bf{98.23} & \bf{98.33}     & \bf{98.47}& \bf{98.65}  \\ \hline

\end{tabu}}
\label{table:main_results}
\end{center}
\end{table}

\begin{table}
\caption{Accuracy results comparison on UCM in percent. MSMatch outperforms all other methods on UCM. The best results per label amount are bold.}
\begin{center}
 {\tabulinesep=1.2mm
\setlength\tabcolsep{2pt}
\begin{tabu} {|X[2.5]|| X[1] | X[1] | X[1]| X[1] |}
 \hline
 & \multicolumn{4}{c|}{Number of labels (per class)}\\
  Work & 100 -- 105 ($\sim 5$) & 200 -- 210 ($\sim 10$) &  400 -- 441 ($20\sim21$) & 1680 \quad (80)\\
 \hline
Dai \etal{} 2019 \cite{dai2019}         & 85.12 & -     & -     & -     \\ \hline
Gu \& Angelov 2018 \cite{gu2018semi}    & 71.86 & -     & -     & -     \\ \hline
Guo  \etal{} 2020 \cite{guo2020}        & 57.10 & 75.69 & 83.33 & 90.48 \\ \hline
Han \etal{} 2018 \cite{han2018}         & -     & 91.42 & 92.68 & -  \\ \hline
Ours                                    & \bf{90.71} & \bf{94.13}  & \bf{95.71}  & \bf{98.33}\\ \hline

\end{tabu}}
\label{table:ucm}
\end{center}
\end{table}

\subsection{Hyperparameters}
Two further factors were investigated in detail. One factor that was found to be particularly critical by Kurakin \etal{}\cite{kurakin2020} is the strength of the weight decay, which penalizes large weights in the neural network. They found $5.0 \cdot 10^{-4}$ to be optimal in most cases. The other one is model size, which is a decisive factor for model performance and can be varied using the different versions (B0 to B7) of the EfficientNet architecture. Note that only B0 to B3 fit into the available GPU memory with the utilized training settings and, thus, no larger models were compared. These runs were only done on EuroSAT due to the large memory requirements of the larger images in UCM. Further, note that results in Tables \ref{table:main_results} and \ref{table:model} were run on different machines, which led to slightly different random seeds and mean values.\\
Results for different weight decay values are displayed in Table \ref{table:weigth_decay}. In our experiments we found that slightly larger weight decay values were beneficial than originally proposed by Kurakin \etal{} \cite{kurakin2020}. The highest accuracy was obtained with a weight decay of $7.5 \cdot 10^{-4}$, which led to an accuracy of 96.63\%.
\\
\begin{table}
\caption{Comparison of different weight decay values in terms of accuracy in percent. All runs are on EuroSAT and used an EfficientNet-B2 and 250 labels. The best result is bold.}
\begin{center}
 {\tabulinesep=0.8mm
 \setlength\tabcolsep{2pt}
\begin{tabu} { |X[1.25m]|| X[1cm] | X[1cm] | X[1cm]| X[1cm] | X[1cm] | X[1cm]| X[1cm] |}
 \hline
  Weight decay & $5.0 \cdot 10^{-5}$ & $7.5 \cdot 10^{-5}$ & $1.0 \cdot 10^{-4}$ & $2.5 \cdot 10^{-4}$ & $5.0 \cdot 10^{-4}$ & $7.5 \cdot 10^{-4}$ & $1.0 \cdot 10^{-3}$\\ \hline
  Accuracy& 94.63 $\pm$0.79 & 93.49 $\pm$1.95 & 95.53 $\pm$1.78 & 95.21 $\pm$1.88 & 95.26 $\pm$3.24& \bt{96.63} $\pm$0.34 & 96.53 $\pm$0.26\\ \hline
\end{tabu}}
\label{table:weigth_decay}
\end{center}
\end{table}
Detailed results for the model size are given in Table \ref{table:model}. We found the EfficientNet-B2 with 9.2 million parameters to provide the best trade-off of performance and model size with an accuracy of 96.85\%. However, performance gains from a larger number of parameters are limited and smaller than standard deviation among random seeds. Thus, choosing a smaller model when optimizing for efficiency can also be reasonable.

\begin{table}
\caption{Comparison of EfficientNets in terms of accuracy and parameters. All models were trained on EuroSAT with 250 labels and a weight decay of $7.5 \cdot 10^{-4}$. The best result is bold.}
\begin{center}
 {\tabulinesep=0.8mm
\setlength\tabcolsep{2pt}
\begin{tabu} { |X[2m]|| X[1cm] | X[1cm] | X[1cm]| X[1cm] |}
 \hline
  Model & B0 & B1 & B2 & B3 \\ \hline
  \# of Model Parameters & 5.3M & 7.8M & 9.2M & 12M \\ \hline
  Accuracy [\%] & 96.78$\pm$0.04  & 96.85$\pm$0.46 & 96.85$\pm$0.56 & \bt{96.91}$\pm$0.27 \\ \hline
\end{tabu}}
\label{table:model}
\end{center}
\end{table}

\begin{figure}[ht]
\centering
\includegraphics[scale=0.25]{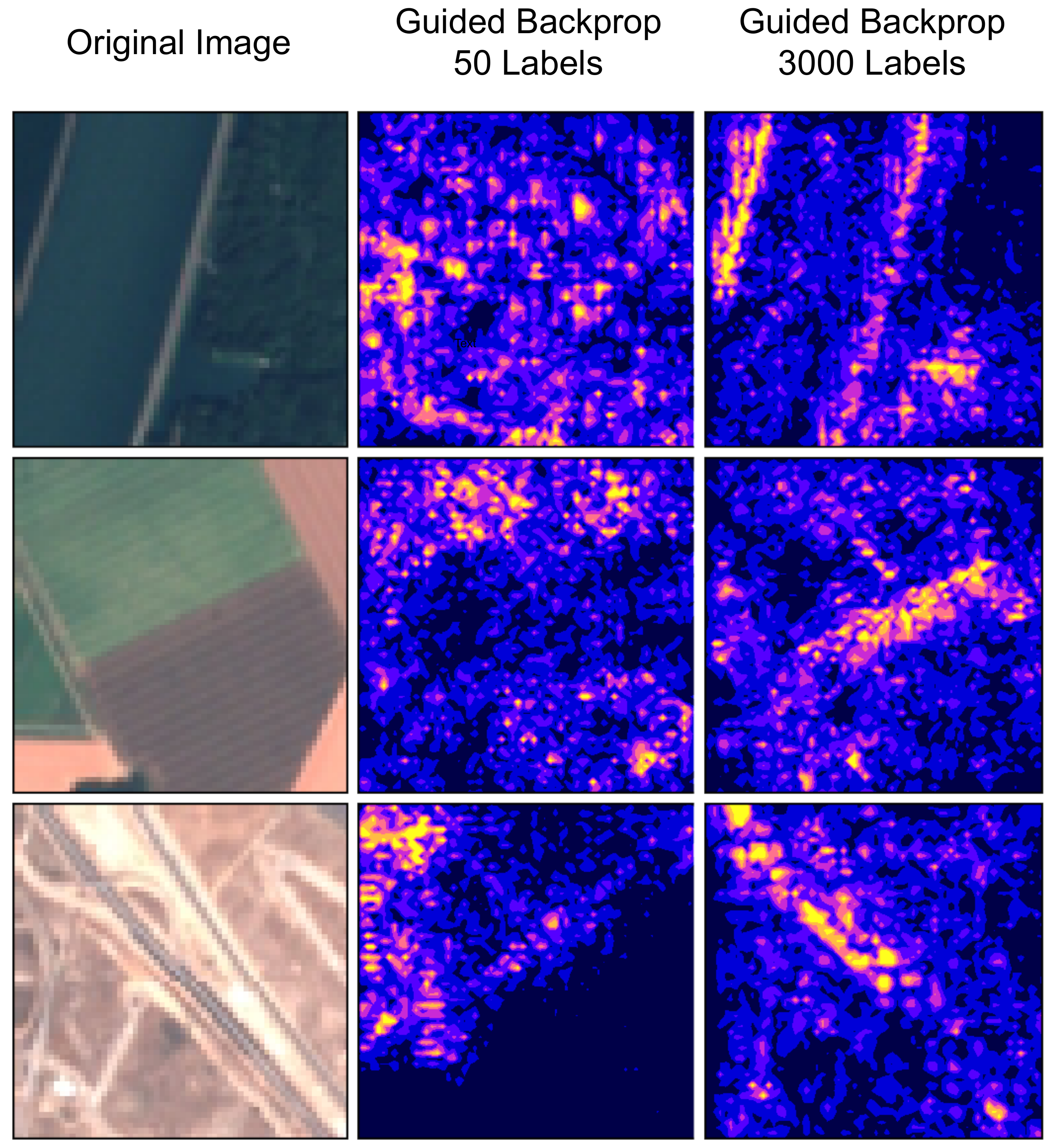}
\caption{Guided backpropagation saliency maps for a model trained on 50 and 3000 labels, respectively. RGB data was used. Images are examples where the 50-label model misclassified the images while the 3000-label model succeeded. The saliency map clearly displays the features that the 3000-label model utilized, which were missed by the 50-label one.}
\label{fig: saliency}
\end{figure}

\begin{table}
\caption{Precision, recall and F1-score metrics for EuroSAT RGB obtained by using an EfficientNet-B2 model trained with five labeled images per class. Results are averaged over three seeds.}
\begin{center}
 {\tabulinesep=0.5mm
\setlength\tabcolsep{2pt}
\begin{tabu} { |X[2]|| X[1c] | X[1c] | X[1c] | X[1c]|}
\hline
Class  & Precision & Recall & F1-score & Support \\ \hline
AnnualCrop             & 94.52  & 97.66 & 96.06 & 300 \\ \hline
Forest                 & 99.55  & 98.22 & 98.88 & 300 \\ \hline
HerbaceousVegetation   & 88.66  & 82.22 & 83.23 & 300 \\ \hline
Highway                & 96.39  & 99.73 & 98.03 & 250 \\ \hline
Industrial             & 97.39  & 98.13 & 97.74 & 250 \\ \hline
Pasture                & 97.21  & 92.83 & 94.97 & 200 \\ \hline
PermanentCrop          & 87.25  & 82.40 & 83.62 & 250 \\ \hline
Residential            & 91.41  & 99.44 & 95.14 & 300 \\ \hline
River                  & 98.49  & 95.46 & 96.95 & 250 \\ \hline
SeaLake                & 99.32  & 98.22 & 98.77 & 300 \\ \hline
\end{tabu}}
\label{table:EuroSAT_RGBMetrics}
\end{center}
\end{table}

\begin{table}
\caption{Precision, recall and F1-score metrics for UCM obtained by using an EfficientNet-B2 model trained with 5 labeled images per class. The support is 20 images for each class. Results are averaged over three seeds.}
\begin{center}
 {\tabulinesep=0.5mm
\setlength\tabcolsep{2pt}
\begin{tabu} { |X[2]|| X[1c] | X[1c] | X[1c]|}
\hline
Class  & Precision & Recall & F1-score \\ \hline
agricultural      & 94.63	 &  88.33	& 91.32     \\ \hline
airplane          & 98.41	 &  100.00	& 99.19     \\ \hline
baseballdiamond   & 93.80	 &  100.00	& 96.79     \\ \hline
beach             & 94.06	 &  100.00	& 96.86       \\ \hline
buildings         & 87.06	 &  75.00	& 79.43     \\ \hline
chaparral         & 93.57	 &  95.00	& 94.27            \\ \hline
denseresidential  & 69.48	 &  33.33	& 44.52     \\ \hline
forest            & 98.41	 &  95.00	& 96.48     \\ \hline
freeway           & 89.29	 &  96.67	& 92.75    \\ \hline
golfcourse        & 87.86	 &  96.67	& 92.02    \\ \hline
harbor            & 95.45	 &  98.33	& 96.83    \\ \hline
intersection      & 100.00	& 100.00	&  100.00     \\ \hline
mediumresidential & 72.05	 &  78.33	& 74.89    \\ \hline
mobilehomepark    & 75.09	 &  93.33	& 83.14     \\ \hline
overpass          & 95.30	 &  98.33	& 96.75       \\ \hline
parkinglot        & 96.97	 &  98.33	& 97.56    \\ \hline
river             & 98.33	 &  86.67	& 91.60    \\ \hline
runway            & 93.31	 &  91.67	& 92.39    \\ \hline
sparseresidential & 96.67	 &  93.33	& 94.91       \\ \hline
storagetanks      & 82.03	 &  90.00	& 85.15      \\ \hline
tenniscourt       & 98.33	 &  96.67	& 97.48      \\ \hline
\end{tabu}}
\label{table:ucmMetrics}
\end{center}
\end{table}

\begin{figure}[ht]
\centering
\includegraphics[scale=0.35]{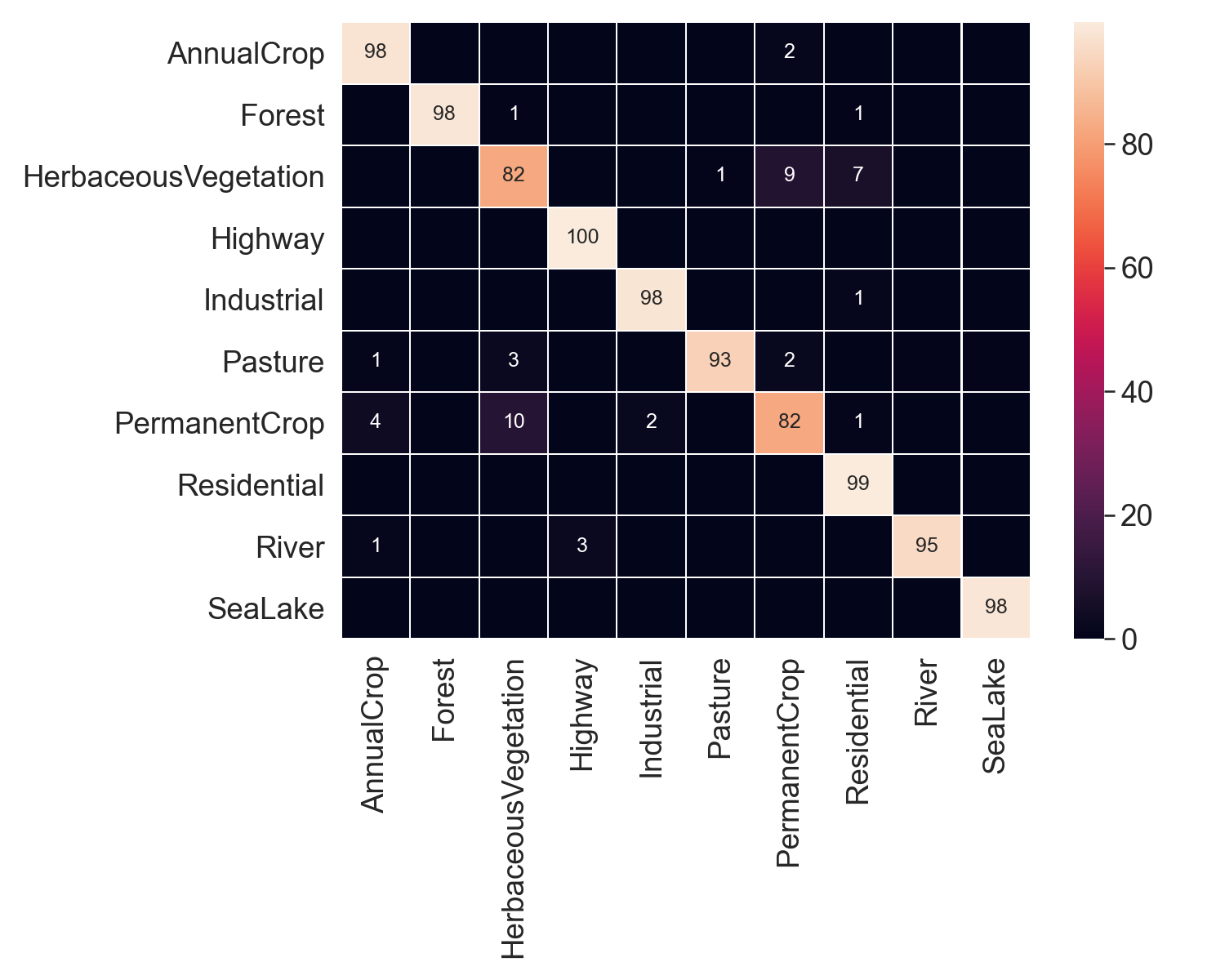}
\caption{Average confusion matrix over three seeds for the EuroSAT RGB dataset obtained through an EfficientNet-B2 model trained with five labeled images per class. Cells containing 0s are not shown. Results are expressed as percentage of the support. Due to rounding numbers may not add up to $1$.}
\label{fig: eurosat_confusion}
\end{figure}

\begin{figure}[ht]
\centering
\includegraphics[scale=0.3]{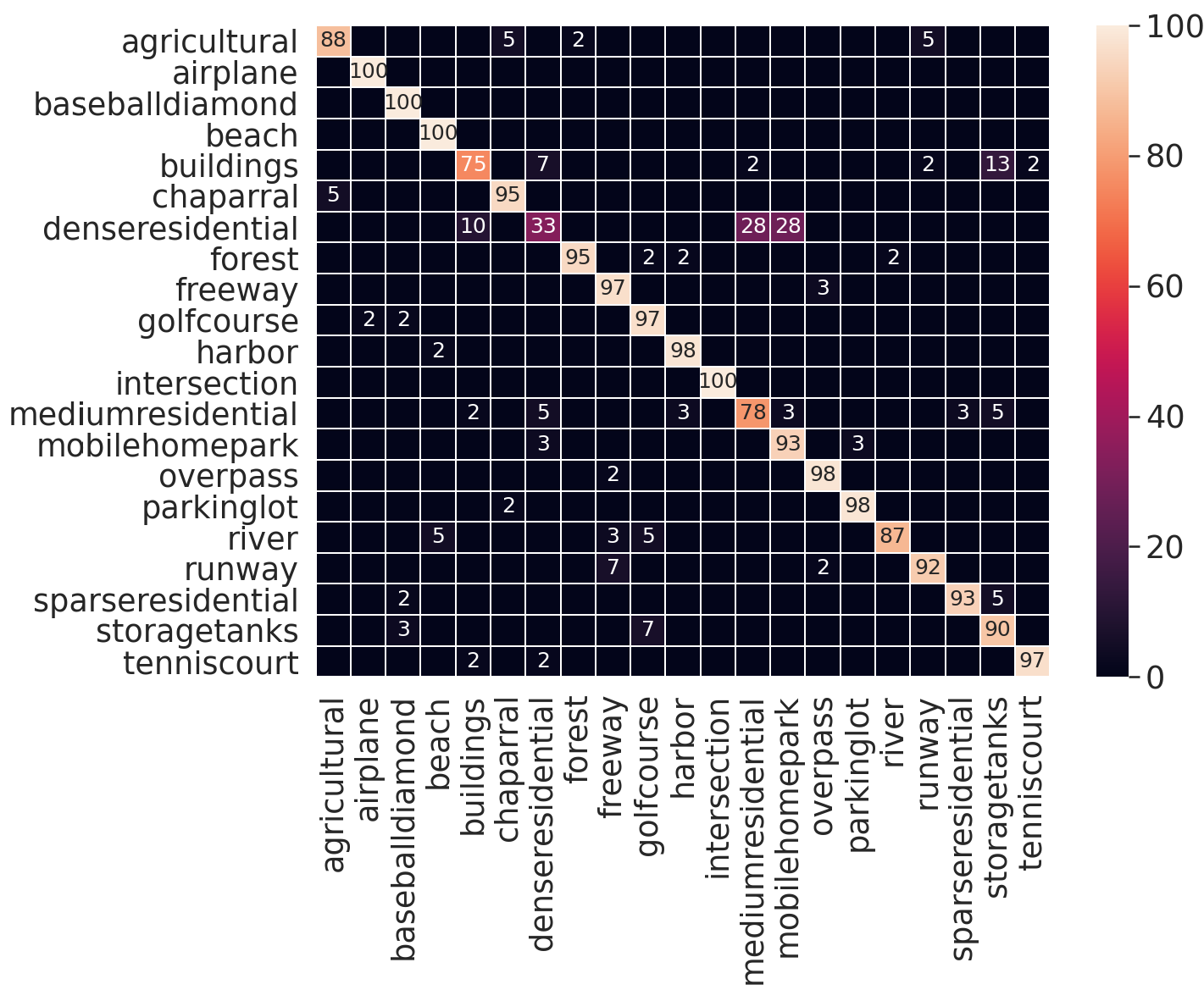}
\caption{Average confusion matrix over three seeds for the UCM dataset obtained through an EfficientNet-B2 model trained with five labeled images per class. Cells containing 0s are not shown. Results are expressed as percentage of the support. Due to rounding numbers may not add up to $1$.}
\label{fig: ucm_confusion}
\end{figure}

\section{Discussion}
Overall, MSMatch outperforms previously published SSL works tested on EuroSAT and UCM. Even with a low number of labels (five per class), it is able to achieve a high accuracy at 94.53\% or 95.86\% for EuroSAT RGB and MS data and 90.71\% for UCM, respectively. This makes the approach applicable in scenarios where only very limited labeled data are available. The superior performance using MS data both highlights the potential of utilizing such data as well as the suitability of the proposed method for it. \\
\subsection{Comparison to FixMatch}
Compared to the original FixMatch approach, we find that slightly higher weight decay values benefit the training. Kurakin \etal{} \cite{kurakin2020} described this for the CIFAR-100 dataset, which features 100 classes. However, EuroSAT, as the other datasets investigated by Kurakin \etal{}, only features ten classes. The benefit of the higher weight decay in our experiments may also be related to the different model choices, as Kurakin \etal{} relied on a different network architecture. \\
In terms of the comparison of model parameters in the network architecture (see Table \ref{table:model}) it is noteworthy that improvements from a larger number of parameters are only marginal. Possibly, the reason is that the EuroSAT dataset features just ten classes. The larger models may be overparameterized for the problem. However, it is also conceivable that the proposed training procedure performs better on smaller models. This will require further investigation in the future. \\
Another element that may warrant more detailed examination in the future is the type of augmentations utilized for the strong augmentation. Kurakin \etal{} \cite{kurakin2020} described some performance impact of the utilized augmentation method. Due to the computational demands of running a large number of test runs the authors were unable to investigate this in more detail. However, it would warrant further investigation, especially in regards to the interplay with MS data.
\subsection{Class-dependent Performance}
Another interesting factor is the varying performance depending on the classes. This tendency is observed in both, UCM and EuroSAT and persists across multiple seeds, i.e. random splits of the train and test sets. A detailed overview of the per class performance for EuroSAT can be seen in Table \ref{table:EuroSAT_RGBMetrics}. Clearly, some classes require more labeled training data than others, which hints at a possible improvement to the described method. In particular, the number of labeled samples could be adjusted for each class in relation to the model's performance on it. For example, after observing the worse performance on the classes \textit{PermanentCrop} and \textit{HerbaceousVegation} in EuroSAT when training with 50 labels (five per class) in an operational scenario, this insight could be used to selectively label more data from such underperforming classes. Note that Figure \ref{fig: eurosat_confusion} highlights that the models tend to confuse these two classes with each other, respectively, and especially the recall of the model on these classes is impacted. In practice, the lower performance for these classes might also hint at an underrepresentation of some necessary features in the supplied training data. This is also evident in Figure \ref{fig: f_scores_eurosat} as the problem is remedied when additional labeled examples for these classes are added. Kurakin \etal{} \cite{kurakin2020} also observed a strong impacted of the selected samples on the performance. In Figure \ref{fig: saliency} we display some saliency maps (using guided backpropagation \cite{uozbulak2019,springenberg2014}), where a model trained on 50 labels (five per class) misclassified the examples whereas a 3000-label model succeeded. Note that the 3000-label model clearly relies on specific contour features in the image that the 50-label model does not recognize. \par
For UCM, the tendency of underperforming in some of the classes was also observed when using 105 labeled images (five images per class). Indeed, as shown in Table \ref{table:ucmMetrics}, the F1-score value obtained by averaging an EfficientNet-B2 model over three seeds is higher than 80\% for all the classes except for \textit{mediumresidential} (74.89\%), \textit{denseresidential} (44.52\%), and \textit{buildings} (79,43\%). In particular, the confusion matrix in Figure \ref{fig: ucm_confusion} shows that misclassified \textit{denseresidential} images are often misidentifed as \textit{mediumresidential} (28\%) and \textit{mobilehomepark} (28\%), and as \textit{buildings} (10\%). This may also indicate that these classes do not feature sufficient distinctive properties for the network to pick up. Similarly, 7\% and 13\% of \textit{buildings} images are predicted as \textit{denseresidential} and \textit{storagetanks}, respectively. This leads to comparatively lower precision on, e.g., \textit{storagetanks} and \textit{mobilehomepark} and low recall on \textit{denseresidential} and \textit{buildings}. Overall, these results highlight that by using five images per class the trained model is not fully capable to distinguish among \textit{mediumresidential},  \textit{denseresidential}, and \textit{buildings} images, where as it can attain robust performance in predicting the other classes. As shown in in Figure \ref{fig: f_scores_ucm} a F1-score higher than 80\% is reached for the \textit{buildings}, \textit{mediumresidential}, and \textit{denseresidential}  classes when training with, respectively, 5, 10, and 21 images per class.

\section{Conclusion}
This works presents MSMatch, a novel SSL approach that is able to vastly improve the state-of-the-art on the EuroSAT and UCM datasets compared to previous works. Depending on the number of labeled training samples, it improves accuracy by between 1.47\% and 18.43\% on EuroSAT and 2.71\% and 7.85\% on UCM compared to previous works. More importantly, it showcases that an accuracy of 95.86\% and 90.71\% on EuroSAT and UCM, respectively, is obtainable with just five labels per class. This bears the promise to make MSMatch applicable to scenarios where a lack of labeled data previously inhibited training neural networks for the task. The method also translates well to MS data, which is, however, harder to process given a lack of GPU-based data augmentation frameworks. \\
Future research will aim to test datasets with even higher resolutions, such as AID, which are computationally more demanding, especially in terms of GPU memory. Adapting MSMatch to a segmentation problem is also conceivable given suitable augmentation methods and might be of interest to broaden the range of possible applications even further. 

\FloatBarrier

\bibliographystyle{IEEEtran}
\bibliography{IEEEabrv,main.bib}


\begin{IEEEbiography}[{\includegraphics[width=1in,height=1.25in,clip,keepaspectratio]{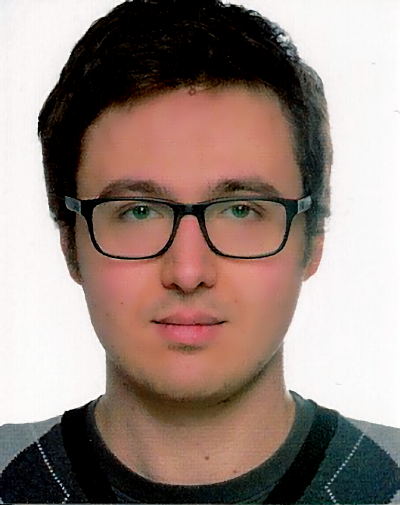}}]{Pablo \gomez}
, PhD, is a research fellow in ESA's Advanced Concepts Team at ESTEC, Noordwijk. He received his PhD from the Friedrich-Alexander-Universit\"at Erlangen-N\"urnberg in 2019 (supervisor Prof. D\"ollinger). He received his M.Sc. in computer science from the Technical University Munich in 2015. Research topics of interest to him range from machine learning and inverse  problems to numerical methods and high performance computing.
\end{IEEEbiography}

\begin{IEEEbiography}[{\includegraphics[width=1in,height=1.25in,clip,keepaspectratio]{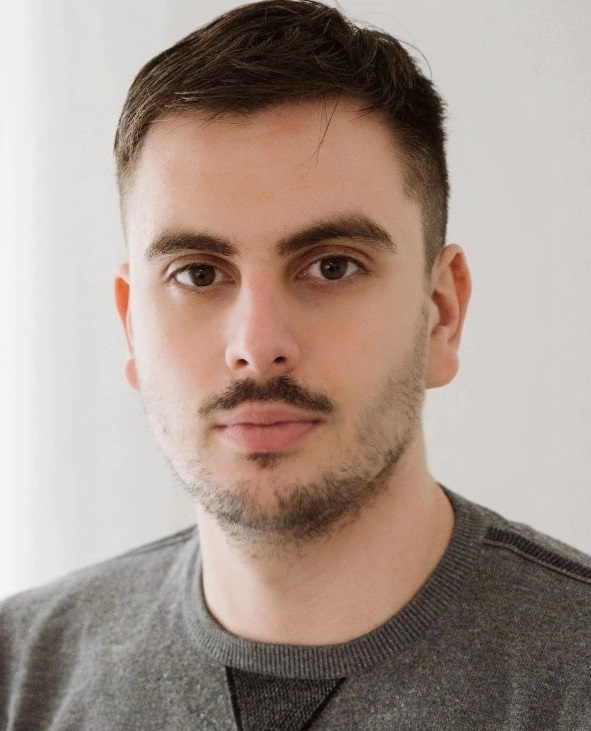}}]{Gabriele Meoni}
, PhD, received the Laurea degree in electronic engineering from the University of Pisa in 2016 and the Ph.D. degree in information engineering in 2020. During his Ph.D., he developed skills in digital and embedded systems design, digital signal processing, and artificial intelligence.  
Since 2020 he is a research fellow in the ESA Advanced Concepts Team. His research interests include machine learning, embedded systems and edge computing.

\end{IEEEbiography}





\end{document}